\documentclass[10pt,twocolumn,letterpaper]{article}

\usepackage{iccv}
\usepackage{times}
\usepackage{epsfig}
\usepackage{graphicx}
\usepackage{amsmath}
\usepackage{amssymb}

\usepackage{amssymb}
\usepackage{pifont}
\usepackage{subcaption}
\usepackage{array}
\usepackage{booktabs}

\usepackage{hyperref}
\hypersetup{pagebackref=true,breaklinks=true,letterpaper=true,colorlinks,bookmarks=false}

\iccvfinalcopy 

\def\iccvPaperID{9} 
\newcommand{\cmark}{\ding{51}}%
\newcommand{\xmark}{\ding{55}}%
\ificcvfinal\fi

\begin{document}

\title{CenterPoly: real-time instance segmentation using bounding polygons}

\author{Hughes Perreault, Guillaume-Alexandre Bilodeau, Nicolas Saunier\\
Polytechnique Montréal\\
Montréal, Canadal\\
{\tt\small \{hughes.perreault, gabilodeau, nicolas.saunier\}@polymtl.ca}
\and
Maguelonne Héritier\\
Genetec\\
Montréal, Canada\\
{\tt\small mheritier@genetec.com}
}

\maketitle
\ificcvfinal\thispagestyle{empty}\fi

\begin{abstract}
We present a novel method, called CenterPoly, for real-time instance segmentation using bounding polygons. We apply it to detect road users in dense urban environments, making it suitable for applications in intelligent transportation systems like automated vehicles. CenterPoly detects objects by their center keypoint while predicting a fixed number of polygon vertices for each object, thus performing detection and segmentation in parallel. Most of the network parameters are shared by the network heads, making it fast and lightweight enough to run at real-time speed. To properly convert mask ground-truth to polygon ground-truth, we designed a vertex selection strategy to facilitate the learning of the polygons. Additionally, to better segment overlapping objects in dense urban scenes, we also train a relative depth branch to determine which instances are closer and which are further, using available weak annotations. We propose several models with different backbones to show the possible speed / accuracy trade-offs. The models were trained and evaluated on Cityscapes, KITTI and IDD and the results are reported on their public benchmark, which are state-of-the-art at real-time speeds. Code is available at \url{https://github.com/hu64/CenterPoly}.
\end{abstract}

\section{Introduction}
%
%
%
%
\begin{figure}[t]
\centering
   \begin{subfigure}[b]{0.4\textwidth}
   \includegraphics[width=\textwidth]{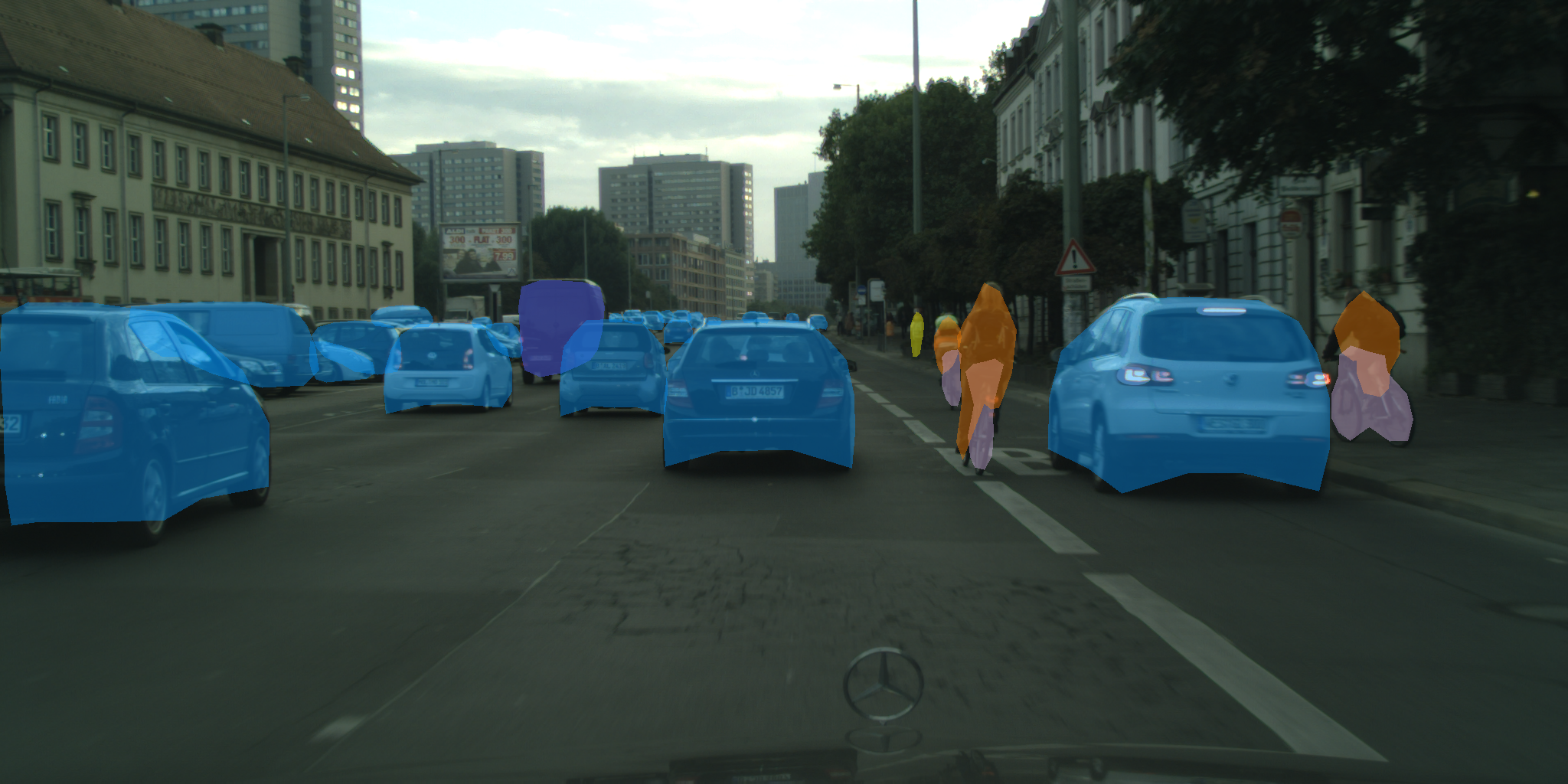}
   \caption{}
   \label{fig:Ng1} 
\end{subfigure}

\begin{subfigure}[b]{0.4\textwidth}
   \fbox{\includegraphics[width=\dimexpr\textwidth-2\fboxsep-2\fboxrule\relax]{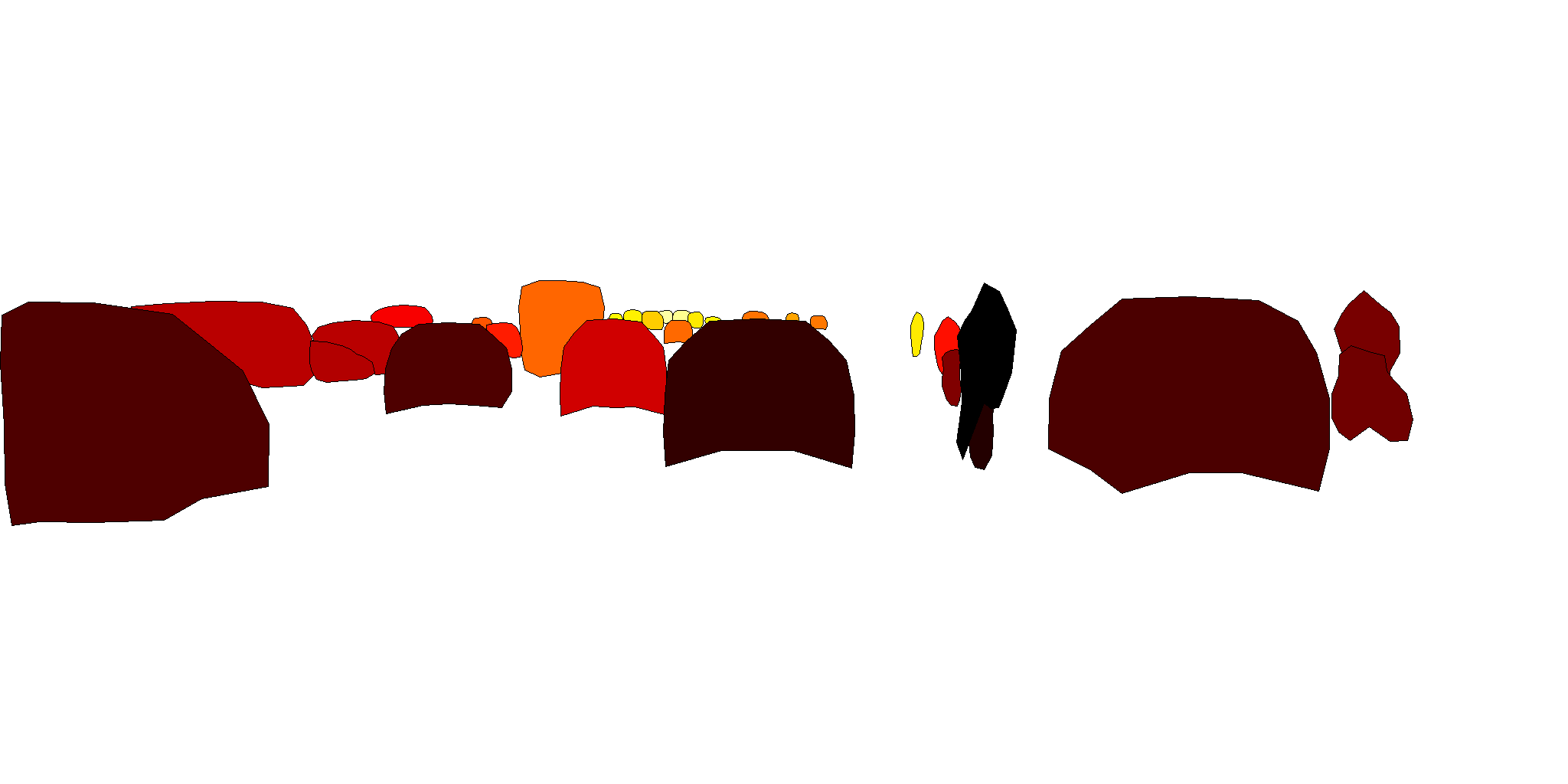}}
   \caption{}
   \label{fig:Ng2}
\end{subfigure}
\caption{CenterPoly produces: (a) an instance segmentation by detecting objects and regressing a bounding polygon for each, and (b) a relative depth value for each object (darker is closer). In this example, polygons have 16 vertices.}
\label{eg1}
\end{figure}


Object detection is typically described as the task of finding a rectangular bounding box around every object of interest in an image, as well as finding a class label for each object. A more complex and demanding task is instance segmentation, where instead of producing a rectangular bounding box, a pixel-wise segmentation is produced for each instance of the objects of interest in the image. As it is more complex, few instance segmentation methods can run as efficiently as object detectors. There is no doubt that real-time instance segmentation (at a frame rate usually faster than 20 fps) is a very useful and an important problem to solve. 

The targeted applications are in the realm of intelligent transportation systems, and they include traffic surveillance, automated driving and other advanced driving assistance systems. These applications often have to deal with dense overlapping objects such as cars that require a segmentation for various tasks such as counting, tracking and re-identification. Having a relative depth map of nearby instances could greatly improve safety in several of these applications to estimate the relative closeness of various nearby road users. Additionally, these applications often require instance segmentation that can be obtained rapidly to take quick decisions about road management or driving, time management being critical in a lot of cases.

Instance segmentation is a harder problem than object detection because it involves an additional task that requires ideally pixel-level precision. Furthermore, adapting an object detector to perform instance segmentation efficiently is not trivial. In object detection, a box can be represented by four values, regardless of the box size. Therefore, this task can be performed quickly. In contrast, an accurate segmentation mask typically requires several hundred values. For example, Mask R-CNN~\cite{he2017maskrcnn} encodes each instance on COCO~\cite{lin2014microsoft} by 14 $\times$ 14 $\times$ 80 = 15680 values. Extracting masks thus requires much more computation. 

For this reason, for keeping our method fast and lightweight, we decided to use bounding polygons to represent the segmentation masks as a compromise on speed and accuracy between bounding boxes and pixel-wise segmentation masks. Polygons can be represented by a few vertices, a parameter which can be adjusted at will. For example, using 12 vertices to represent an object (i.e. 24~numbers) can produce a much more precise segmentation than a bounding box at a much lower cost than using masks. But using polygons also brings limitations, among others the fact that a polygon cannot have holes or be fragmented and the fact that the segmentation accuracy is limited by the number of vertices. This is partly addressed with a relative depth map that places an object instance polygon in front or behind others. 

We approach instance segmentation by modifying a fast anchor-free object detector, CenterNet~\cite{centernet_zhou2019objects}, to produce bounding polygons for each detected object, using a fixed number of vertices for each polygon. The resulting method, called CenterPoly, is almost as fast as CenterNet and the resulting masks are accurate compared to other real-time instance segmentation methods. In figure~\ref{eg1}, we show an example of the result of our method.

One of the weaknesses of bounding polygons compared to masks is the object overlapping problem. When two polygons overlap, there is an ambiguity regarding which polygon the overlapping pixels belong to. To solve this problem, we added a head to our network that learns the relative depth of objects. The relative depth is sometimes available in the annotations, and we used a transfer learning strategy when it is not. We discuss this further in section~\ref{relative-depth}. This head produces one value per pixel, and this value reflects the relative depth of the object with its center at this location. 

In summary, we propose CenterPoly, a novel real-time instance segmentation method evaluated on the Cityscapes~\cite{cordts2016Cityscapes}, KITTI~\cite{Alhaija2018IJCV_kitti} and IDD~\cite{varma2019idd} datasets. The contributions are as follows:
\begin{itemize}
        \item We designed a network head to produce bounding polygons for every detected object in an image.
        \item To produce accurate ground-truth polygon vertices, we designed a vertex selection policy, which proves to be very important for polygon regression. 
        \item To solve the issue of overlapping instances, we trained the network to learn the relative depth of objects. 
        \item We proposed two modifications to the CenterNet center heatmaps to better suit instance segmentation. 
\end{itemize}

\section{Related Work}
As we use a segmentation by detection paradigm, we highlight the most important work in this field in the last few years. 
\subsection{Object detection}
\textbf{Two-stage detectors} have a first phase where they find object candidates and a second phase where they refine and choose the best of those candidates. It is the slowest object detection paradigm. The most notable two-stage detector is certainly the ground-breaking Faster R-CNN~\cite{ren2015faster}, which is the third iteration of the R-CNN~\cite{rcnn_Girshick_2014_CVPR} family. Faster R-CNN uses a shared backbone between two networks, a region proposal network (RPN) and a refinement network. The RPN proposes the best candidate bounding boxes by using anchor boxes of different predefined sizes and aspect ratios. The refinement network then classifies and refines the bounding box candidates, and keeps the ones with the highest scores. 

\textbf{One-stage detectors} remove the object candidate search phase, and rather try to find objects at the same time as classifying them and refining their position and size. As a result, they are much faster than two-stage detectors. YOLO~\cite{yolo_redmon2016you} was the first method to use this paradigm. It consists of a network to detect objects based on a division of the image into a regular grid, and having each cell predict a certain number of objects close to them. The two subsequent iterations of the method, YOLO9000~\cite{redmon2017yolo9000} and YOLOv3~\cite{redmon2018yolov3}, improved upon it by switching to anchor boxes and designing a better and deeper backbone network among others. SSD~\cite{liu2016ssd} introduced a scheme to merge feature maps at different resolutions before applying anchor boxes, and thus tackles the problem of detecting objects at different scales. RetinaNet~\cite{lin2018focal_retina} uses an architecture that is fairly similar to SSD, although a feature pyramid network~\cite{Lin_2017_CVPR_FPN} is used for multi-scale detection. They also introduce the focal loss, which is a loss designed to help counter the imbalance between positive and negative examples. Chen \etal~\cite{chen2021edge} proposed a modified one-stage detector based on YOLOv3 for fast and low memory traffic flow detection. They modified YOLO's backbone by adding dense connections to it instead of the residual modules, and by lowering the spatial resolution with max-pooling. 

\textbf{Anchor-free detectors} refer to the relatively newer paradigm of detection by object keypoints. They can also be classified as one-stage detectors. CornerNet~\cite{law2018cornernet} trains a network to recognize top-left and bottom-right corners, using corner pooling. It also has an embedding branch so that if the top-left corner and the bottom-right corner belong to the same object, their embedding will be similar, and different otherwise. They use this embedding to pair corners. Keypoint Triplets~\cite{duan2019centernet} improves upon CornerNet by adding its center as the third keypoint. They use this center keypoint and its confidence score to remove false positives by testing each bounding box (paired corners) for a confident center keypoint at its center. They also improve the corner pooling layers. The ``objects as points'' approach~\cite{centernet_zhou2019objects} presents a surprisingly simple architecture, detecting objects as their center keypoint, as well as their height and width. SpotNet~\cite{perreault2020spotnet} builds upon objects as points by introducing a self-attention module trained with semi-supervised annotations.

\subsection{Instance Segmentation}
The natural progression from bounding box detection is a finer pixel-by-pixel segmentation. Mask R-CNN~\cite{he2017maskrcnn} builds upon Faster R-CNN by adding a size invariant mask branch in the second phase. For each object candidate, it produces one mask for each possible label. The same improvement exists for RetinaNet, named RetinaMask~\cite{fu2019retinamask}. In RetinaMask, a mask subnetwork is added to produce an instance mask for each object candidate. PANet~\cite{liu2018path_panet}, a variant of Mask R-CNN, produces accurate masks adding a bottom-up path augmentation to an FPN and adaptive feature pooling.  
 
Some methods perform instance segmentation by detecting a bounding polygon around objects. PolarMask~\cite{xie2020polarmask} and Poly-YOLO~\cite{hurtik2020poly-yolo} use a polar grid to represent vertices by their angles from the center. The main difference between them is that Poly-YOLO learns size invariant shapes using a normalized distance. Although our method share some similarities with them, there also are fundamental differences in the vertices representation, in the training as well as the addition of a the novel depth branch. Closer to our proposed method, Polygon-RNN++~\cite{acuna2018efficientpolygonrnn} uses a CNN and RNN architecture to detect one polygon vertex at a time, and is therefore rather slow although quite accurate. 

A few methods can produce instance segmentation in real-time, including Poly-YOLO, but these works remain scarce. Box2Pix~\cite{UB18} can produce results on Cityscapes at 10.9~fps by combining bounding box detection as well as semantic segmentation, and predicting pixel offset to the object centers. The spatial sampling network method~\cite{Mazzini_2019_CVPR_Workshops} can achieve very fast instance segmentation on Cityscapes at 113 fps using a decoder network and thresholding at inference time.  Yolact~\cite{bolya2019yolact} works by producing a few segmentation prototypes and trains a network to learn the proper coefficient to combine them. It then uses bounding box detection to crop a region in the combined prototypes. ESE-Seg~\cite{eseseg_xu2019explicit} performs detection and segmentation simultaneously by encoding instance as a novel shape representation called Inner-center radius that allows the method to select useful contour points. SOLO~\cite{wang2020solo} and its follow-up SOLOv2~\cite{wang2020solov2} are fast instance segmentation methods that perform location classification while treating instances as categories. SOLOv2 improves upon it by splitting the network into two branches, one that generates mask kernels and another one that produces the feature maps that are convolved. Finally, Deep Snake~\cite{peng2020deepsnake} is a fast instance segmentation method that learns to deform the generated contour of an object to iteratively better match the contour of the object. The method works in two stages, contour generation followed by contour deformation. 

\section{Proposed Method}
\begin{figure*}[ht]
\begin{center}
\includegraphics[width=0.9\linewidth]{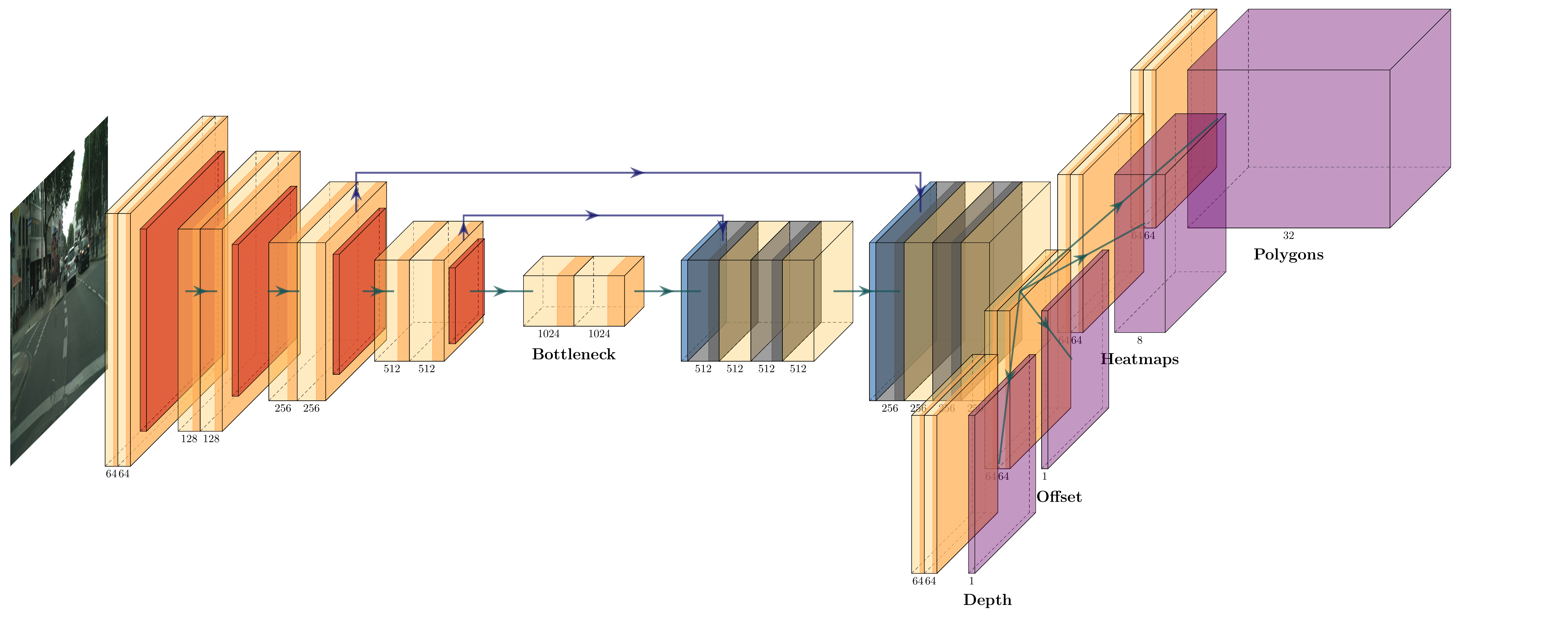}
\end{center}
\caption{An overview of the CenterPoly architecture. The image first passes through a CNN backbone, displayed here as an Hourglass network. The feature map is then shared between four network heads, the polygon regression head, the center heatmaps head used for detections, the object offset head and the relative depth map head. The sizes displayed are for illustration purposes only, please refer to the code for the detailed architecture. }
\label{centerpoly}
\end{figure*}

CenterPoly is based on the CenterNet~\cite{centernet_zhou2019objects} object detector. CenterNet detects objects using three heads, one to produce heatmaps of the center of objects for each label, one to regress their width and height and one to regress the object offsets. We propose to replace the width and height head by a polygon regression head, which will regress a fixed number of points around the object representing a bounding polygon. We modify the heatmaps to account for object proportions and to better deal with instance segmentation. We also add a head to model relative depth to find which objects are further or closer in the scene. This helps in segmenting overlapping objects. This information is available implicitly in some dataset annotations. Indeed, the relative depth may be available if the annotators always annotated objects that are behind others before those in front, as it is the case in Cityscapes~\cite{cordts2016Cityscapes} and IDD~\cite{varma2019idd}. To train on datasets where this information is not available, it is possible to use transfer learning. We use this order of appearance of objects in the annotations as relative depth ground-truth. It is by no means an absolute depth or even a ``full'' annotation, as two objects with the same depth will have different relative depth value. Therefore, this can be considered as a weak annotation. Despite this, it has proven to be sufficient for our purposes, as we will see in the results. An overview of our model can be seen in figure~\ref{centerpoly}.

\subsection{Modified Heatmaps}
CenterNet uses ground-truth center heatmaps that are shaped like circular Gaussians to train the network. These kinds of heatmaps will penalize similarly the same error on the vertical axis and on the horizontal axis, even for thin rectangular boxes, where an error is more costly on the box short axis. To help overcome this problem, we are instead using elliptical Gaussians that use the ratio of the rectangular bounding box to get the ratio of the small and the large radii of the ellipse. Our small radius is the same as the circular radius from CenterNet, while our large radius captures the object proportions to penalize center detection errors more accurately. 

The use of instance segmentation ground-truth has allowed us to make another tweak to the CenterNet heatmaps. Using the center of the bounding box as the object center can be misleading, as the object might be distributed unevenly in the bounding box (for instance, a pedestrian with an arm up). As an alternative, we use the object center of gravity by computing the mean of the vertex locations on the object contour. This gives us ground-truth centers that are more adapted for computing polygon vertices offsets, as we will see in the next section. 

\subsection{Polygon Regression Head}
Our polygon regression head is composed of one $3 \times 3$ convolution followed by a $1 \times 1$ convolution to reduce the dimension, and outputs $N \times 2$ floats at each location on the spatially downsampled feature map, $N$ being the number of vertices used to define each polygon. The values represent offsets from the object center, paired in $x$ and $y$ coordinates. The points are arranged in such a fashion that the first one is always the one on a line that goes from the center to the top-left corner, and the subsequent ones are in clockwise order. For instance, if we use a model with $N$ vertices, the output of the polygon head at location $(x, y)$ would be ($i_{1}, j_{1}, i_{2}, j_{2}, ...,  i_{N}, j_{N})$ with $(x + i_{n}, y + j_{n})$ being the coordinates of the $n^{th}$ polygon vertex. 

Although this head produces dense polygon predictions, the polygons used for training are only the ones with associated objects. This means that predictions at locations where there are no object centers will not contribute to the loss. To achieve that, we use a maximum number of objects per image and a ground-truth mask to hide irrelevant predictions. We train this head using a standard $L1$ loss, that is:
\begin{equation}
L_{poly} =  L1Loss(poly, poly_{GT}), 
\end{equation}
where $poly$ represents the regressed polygons with associated objects and $poly_{GT}$ is the ground-truth.

\subsection{Vertex Selection Policy}
\label{policy}
\begin{figure}[ht]
\begin{center}
\includegraphics[width=1\linewidth]{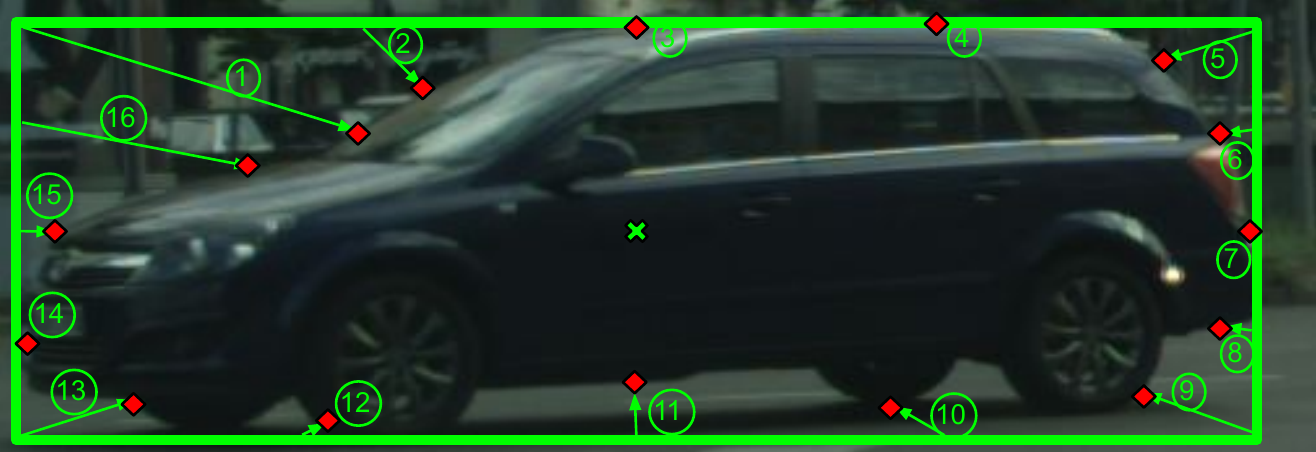}
\end{center}
\caption{The vertex selection strategy: we trace lines at regular intervals in the bounding box, starting at the top left corner and going clockwise. The selected vertex is the first point on the line within the instance mask when going from the bounding box toward the center. The interval changes depending on the number of vertices used in the method.}
\label{vertices}
\end{figure}
During our experiments, the vertex selection was found to be crucial to train the model to learn the shapes of the polygons. The initial annotations come in two formats, images with label ids for each pixel, as well as bounding polygons (which are not as precise due to overlapping, fragmentation, etc.). Neither of those formats are suitable for us, the images with ids for obvious reasons and the polygons because the number of vertices is not fixed for each instance, and the density of points around the objects is far from uniform in most cases, meaning that more points could be on one side of the object. 

The first strategy was to create polygons with a fixed number of vertices using the ground-truth bounding polygons, and either adding vertices between distant ones or removing vertices between close ones. However, the network had a really hard time learning these vertices because their positions varied too much on the object boundaries and could not produce good results.

The adopted strategy, shown in figure~\ref{vertices}, was designed so that the network can better model what each vertex represents, so that each $n^{th}$ vertex represents something similar for each instance. First, we calculate the tightest bounding box surrounding the instance mask. We divide a number of points equally between the four sides of the bounding box. Then, at these points on each side of the bounding boxes, we trace a line between the point on the box and the center of the box. We find the first point that is non-zero on the instance mask, and add it to our ground-truth vertices. We then step on the following point along the sides of the bounding box and continue the process, until we reach the starting point. As a result of this process, each $n^{th}$ vertex have approximately the same angle for all object instances, which facilitates the learning. 

\subsection{Relative Depth Head}
\label{relative-depth}
Our method finds a bounding polygon for each instance in the image. However, this is not sufficient to achieve the best performance, as a pixel can only be associated with one instance in the ground-truth. Therefore, for overlapping instances, we must decide which instance goes in front of which. In the Cityscapes and the IDD annotations, the objects were recorded in order of distance, from furthest to the closest (generally). We use this relative depth as ground-truth to train a dense relative depth head that is trained over the whole image so that an object with the center $(x, y)$ will have depth $D(x, y)$, where $D$ is our two-dimensional depth map. To train this branch, we use the same strategy as for the polygon regression branch, where only the values with associated objects will contribute to the loss. The training loss is:

\begin{equation}
L_{depth} =  L1Loss(depth, depth_{GT}) 
\end{equation}
where $depth$ represents the regressed depth values with associated objects and $depth_{GT}$ is the ground-truth.

The architecture of our relative depth head is the same as the polygon regression head, specifically one $3 \times 3$ convolution followed by a $1 \times 1$ convolution to reduce the dimension: the head outputs one float at each location on the spatially downsampled feature map, the relative depth value. The smaller the value, the further the object. We use this relative depth value to place instances in front of other instances for high confidence instances only. As a result, an instance with a confidence score lower than a threshold will never hide another object, although the instance is still included in the results for evaluation. We can visualize in figure~\ref{qualitative} the kind of relative depth map obtained in the form of heatmaps. Please note that the depth is relative from one instance to another and is not an absolute measure of the pixel depth, as absolute depth information is not available. 


\subsection{Training}
Our ground-truth vertices are produced beforehand, as described in section~\ref{policy}, and used for training our model. We train our polygon and relative depth head from scratch, and use pre-trained weights from MS-COCO~\cite{lin2014microsoft} for the center heatmaps and the offset heads. Our heads are integrated in CenterNet and the model is trained end-to-end. Our loss function is as follows: 
\begin{multline}
Loss = W_{hm} * L_{hm} + W_{poly} *L_{poly} + \\ W_{depth} * L_{depth} + W_{offset} * L_{offset}, 
\end{multline}
where the heatmap loss $L_{hm}$ (focal loss) and the offset loss $L_{offset}$ ($L1$ loss) are the same as the ones in CenterNet. $W_{hm}$, $W_{poly}$, $W_{depth}$ and $W_{offset}$ are the respective weights. We talk in depth about the implementation details in section \ref{implemenation}. 

\section{Experiments}
We trained and evaluated our model on the Cityscapes, KITTI and IDD datasets for the instance segmentation task and report our results as they appear on their public benchmarks. An ablation study was also performed on Cityscapes. 

\subsection{Evaluation Datasets}
Cityscapes~\cite{cordts2016Cityscapes} is a dataset of street scenes captured in several cities in Germany. The image resolution is 2048 $\times$ 1024. A set of 8 image labels is defined which includes person, rider, car, truck, bus, train, motorcycle and bicycle. The other labels are ignored during the instance segmentation evaluation. Three image sets are predefined as train, validation and test. The provided annotations are only for the train and validation sets. The AP (average precision) used by the Cityscapes dataset is the MS-COCO AP, defined as AP[50:95] with steps of 0.05, which is the average of the AP values with minimum IOU (intersection over union) ranging from 0.50 to 0.95. Cityscapes also computes the AP for objects limited to a 50-m range and a 100-m range. KITTI Vision~\cite{Alhaija2018IJCV_kitti} for instance segmentation is a small dataset that consists of 200 train and 200 test images of street scenes with the same labels as Cityscapes. The image resolution is approximately 1280 $\times$ 384. KITTI uses the AP and the AP50 as defined above. The India Driving Dataset (IDD)~\cite{varma2019idd} consists of street scene images captured in India. The image resolutions vary between 1920$\times$ 1080 and  1280$\times$ 964. The train/val/test split is approximately 7000/1000/2000 images respectively. This dataset also uses the AP and the AP50 as defined above.

\subsection{Implementation Details}
\label{implemenation}
We implemented CenterPoly with Pytorch~\cite{NEURIPS2019_9015_pytorch} and trained it for 240~epochs on a single GTX 1080 Ti using the adam optimizer~\cite{kingma2014adam}. Our inference speeds are shown for this GPU, which is not the fastest on the market by far. For our three tested backbones, DCNv2 ResNet-18, DCNv2 ResNet-50~\cite{he2016deep_resnet} and the hourglass~\cite{newell2016stacked}, we used CenterNet weights pre-trained on MS-COCO. We then added the Polygon regression and depth map heads and continued the training on Cityscapes. For IDD and KITTI, we started the training after training on Cityscapes. The loss weights are $W_{hm} = 1$, $W_{poly} = 1$, $W_{depth} = 0.1$ and $W_{offset} = 0.1$. As we do not have any depth annotations for KITTI, we use fake depth annotations and set $W_{depth} = 0$. We use a batch size of six, a learning rate of 2e-4 and drop the learning rate by a factor of ten at epoch 90 and 120 on each tested dataset. For the evaluation, we only consider objects with a confidence score of over $0.5$ for the depth map, which means no object with a confidence below that is allowed to hide another one. We trained at a resolution of 1024$\times$512 as our GPU memory is limited. During training, we used standard data augmentation techniques such as random cropping and flipping. For more details on the implementation, please refer to our code. 

The backbone used for CenterPoly main results is the Hourglass Network~\cite{newell2016stacked}, as it proved to be very efficient for CenterNet and keypoint detection in general. The encoder-decoder architecture of the Hourglass has a very high expressivity that is useful for the multi-task training that CenterPoly does. However, in our efforts to reach real-time speeds, we reduced the number of stacks to only one compared to two in CenterNet. Also, we used 16 polygon vertices to segment instances (For how we chose this number, see section~\ref{secablation}). 

\subsection{Results}

\begin{table*}
\begin{center}
\caption{Results on the Cityscapes, KITTI and IDD test sets, as shown on their respective public benchmarks, order by increasing speed from top to bottom. When not available, runtimes were estimated. Mask type: Full: based on pixel-wise labels, polygon: based on a bounding polygon. Boldface: best results. Results for PANet and Mask R-CNN on IDD were taken from the original IDD paper~\cite{varma2019idd}.}
\begin{tabular}{|l|l|c|c|c|c|c|}
\hline
Method & Mask type & AP & AP50\% & AP100m & AP50m & Runtime(s) \\
\hline\hline
\multicolumn{7}{c}{\textbf{Results on Cityscapes}} \\
\hline\hline
PANet~\cite{liu2018path_panet} & Full & \textbf{31.80} & \textbf{57.10} & \textbf{44.20} & \textbf{46.00} & - \\
Mask R-CNN~\cite{he2017maskrcnn} & Full & 26.22 & 49.89 & 37.63 & 40.11 & \textbf{0.2} \\
LCIS~\cite{hsu2018learningLCIS} & Full & 15.10 & 30.80 & 24.20 &  25.80 & $>$ 0.2 \\
InstanceCut~\cite{kirillov2017instancecut} & Full & 13.00 & 27.90 & 22.10 &  26.10 & - \\
Recurrent Attention~\cite{ren2017endrecattend} & Full & 9.50 & 18.90 & 16.80 &  20.90 & $>$ 0.33 \\
\hline
CenterPoly (Ours) & Polygon &\textbf{15.54} & \textbf{39.49} & \textbf{23.33} & \textbf{24.45} & 0.046 \\
Box2Pix~\cite{UB18} & Full & 13.10 & 27.20 & - &  - & 0.09 \\
Spatial Sampling Net~\cite{Mazzini_2019_CVPR_Workshops} & Full & 9.20 & 16.80 & 16.40 & 21.40 & \textbf{0.009} \\
Poly-YOLO~\cite{hurtik2020poly-yolo} & Polygon & 8.70 & 24.00 & - & - & 0.046 \\
Poly-YOLO lite~\cite{hurtik2020poly-yolo} & Polygon & 7.80 & 21.70 & - & - & 0.026 \\
\hline

\hline\hline
\multicolumn{7}{c}{\textbf{Results on KITTI}} \\
\hline\hline
UniDet\_RVC~\cite{zhou2021simple_unidet} & Full & \textbf{23.19} & \textbf{49.13} & - & - &  \textbf{0.3} \\
BAMRCNN\_ROB & Full & 0.68 & 1.81 & - & -  & 1 \\
\hline
CenterPoly (Ours) & Polygon & \textbf{8.73} & \textbf{	26.74} & - & -  & \textbf{0.046} \\
\hline

\hline\hline
\multicolumn{7}{c}{\textbf{Results on IDD}} \\
\hline\hline
PANet~\cite{liu2018path_panet} & Full & \textbf{37.60} & \textbf{66.10}& - & -  & - \\
Mask R-CNN~\cite{he2017maskrcnn} & Full & 26.80 & 49.90 && - & \textbf{0.2}\\
\hline
CenterPoly (Ours) & Polygon & \textbf{14.40} & \textbf{36.90}& - & -  & 0.046\\
Poly-YOLO~\cite{hurtik2020poly-yolo} & Polygon & 11.50 & 26.70& - & -  & 0.049\\
Poly-YOLO lite~\cite{hurtik2020poly-yolo} & Polygon & 10.10 & 23.90& - & -  & \textbf{0.027}  \\

\hline
\end{tabular}
\label{results}
\end{center}
\end{table*}

Our detailed results on the three datasets are presented in table~\ref{results}. We compare CenterPoly to other fast methods that have been tested on each dataset. We also added some slower methods to compare as baselines. Even though we outperform every real-time method with the AP metric, CenterPoly outperforms the other real-time methods by an even larger margin for the AP50\% metric. For instance on Cityscapes, if for the AP, we outperform LCIS by only 0.44, CenterPoly outperforms it for the AP50\% metric by 8.69. This shows that our masks are very accurate for a coarse segmentation, but slightly less for a very fine segmentation, which can be explained by the nature of polygons. On KITTI, although no other real-time instance segmentation method have submitted results, CenterPoly still presents good results for real-time instance segmentation and presents a baseline for future methods to compare to. On the IDD dataset, we compare CenterPoly to other methods that have published results on it, and presents state-of-the-art results at real-time speed.

\begin{figure*}[t]
        \vspace{-1.2em}

        \centering
        
        \begin{subfigure}[b]{0.3\textwidth}
            \centering
            \includegraphics[width=\textwidth]{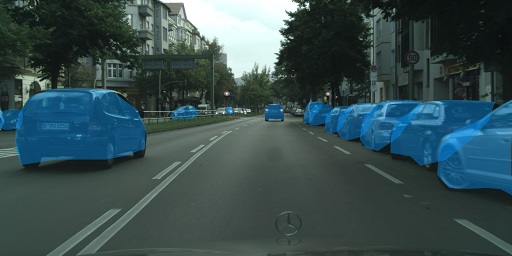}
        \end{subfigure}
        \hspace{0.5em}
        \begin{subfigure}[b]{0.3\textwidth}  
            \centering 
            \includegraphics[width=\textwidth]{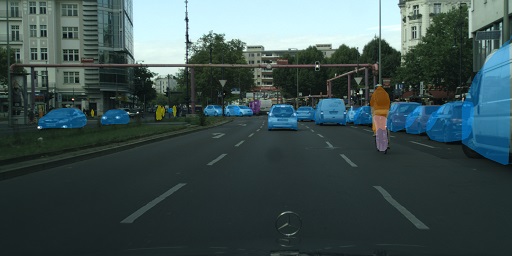}
        \end{subfigure}
        \hspace{0.5em}
        \begin{subfigure}[b]{0.3\textwidth}
            \centering
            \includegraphics[width=\textwidth]{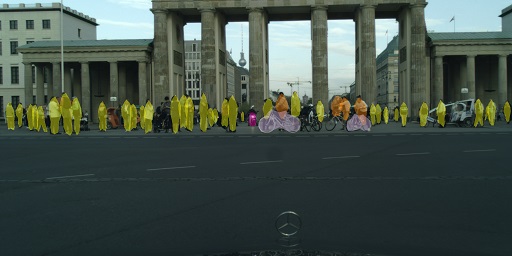}
        \end{subfigure}
        \vskip\baselineskip
        \begin{subfigure}[b]{0.3\textwidth}
            \centering
            \fbox{\includegraphics[width=\dimexpr\textwidth-2\fboxsep-2\fboxrule\relax]{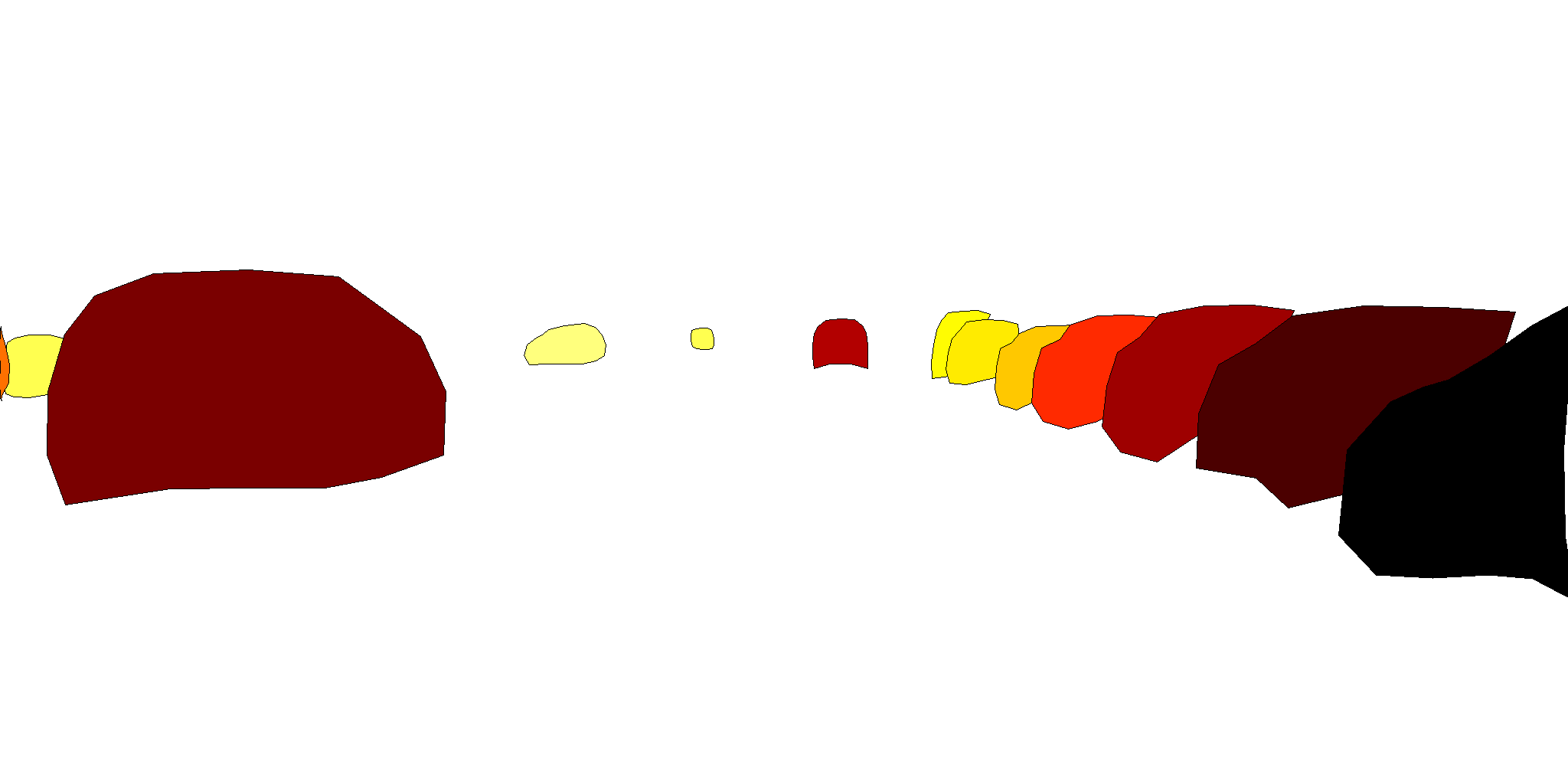}}
        \end{subfigure}
        \hspace{0.5em}
        \begin{subfigure}[b]{0.3\textwidth}  
            \centering 
            \fbox{\includegraphics[width=\dimexpr\textwidth-2\fboxsep-2\fboxrule\relax]{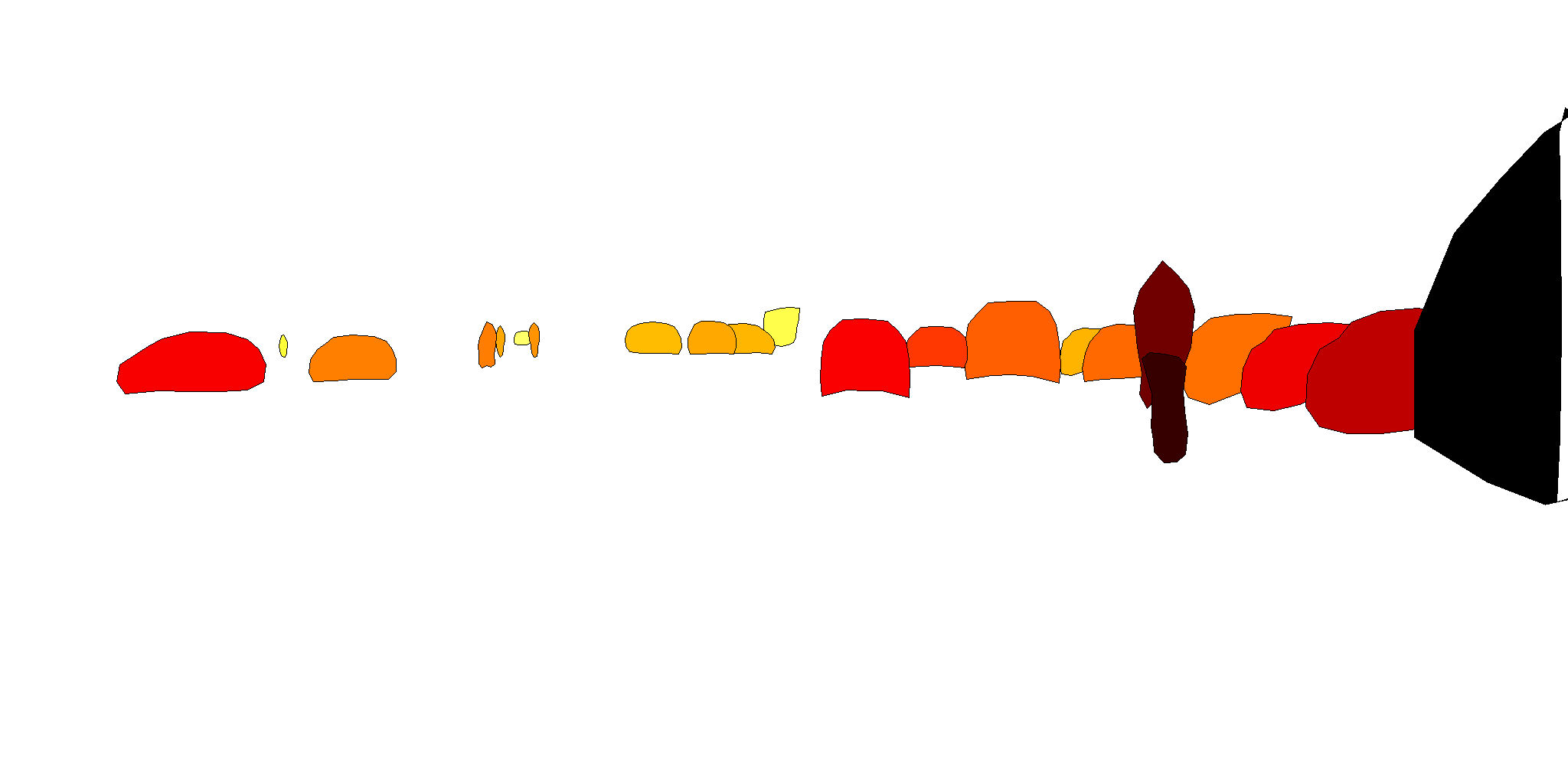}}
        \end{subfigure}
        \hspace{0.5em}
        \begin{subfigure}[b]{0.3\textwidth}
            \centering
            \fbox{\includegraphics[width=\dimexpr\textwidth-2\fboxsep-2\fboxrule\relax]{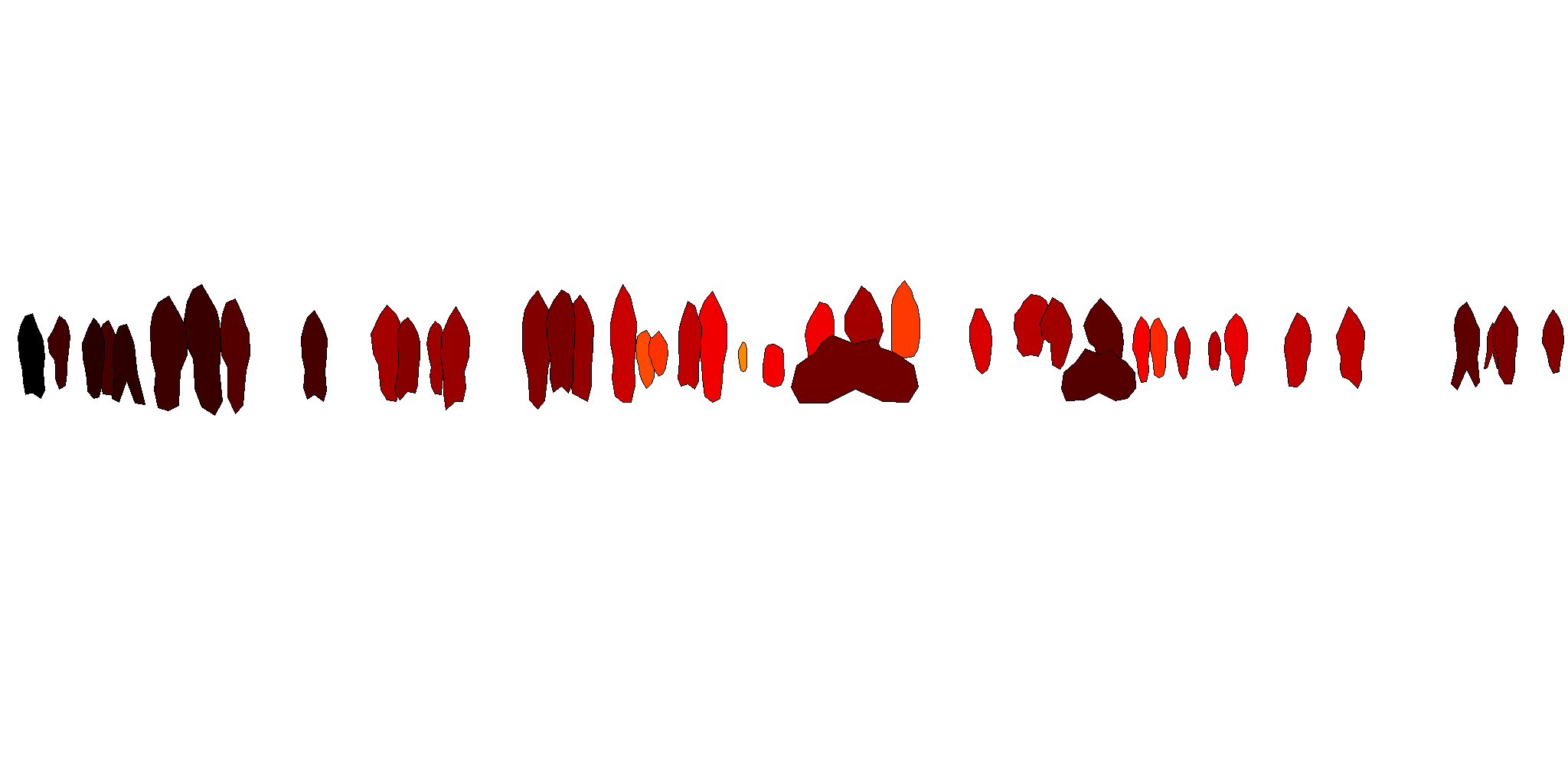}}
        \end{subfigure}
        \vskip\baselineskip
        \begin{subfigure}[b]{0.3\textwidth}
            \centering
            \includegraphics[width=\textwidth]{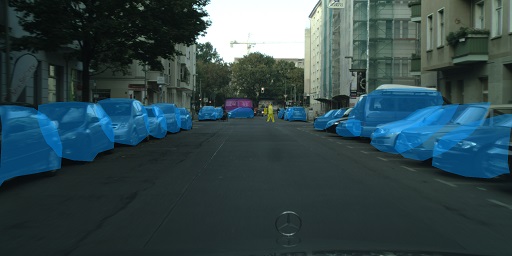}
        \end{subfigure}
        \hspace{0.5em}
        \begin{subfigure}[b]{0.3\textwidth}  
            \centering 
            \includegraphics[width=\textwidth]{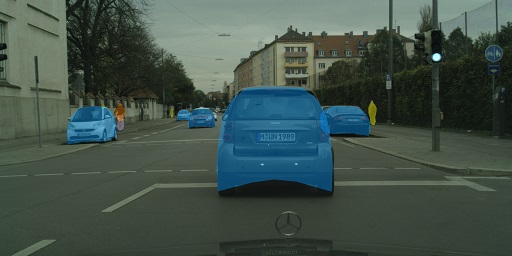}
        \end{subfigure}
        \hspace{0.5em}
        \begin{subfigure}[b]{0.3\textwidth}
            \centering
            \includegraphics[width=\textwidth]{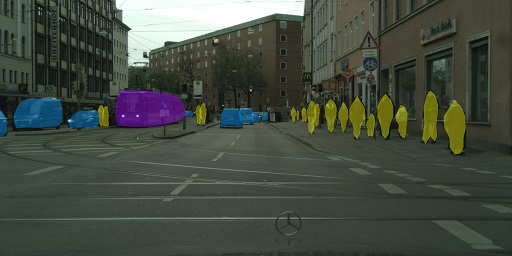}
        \end{subfigure}
        \vskip\baselineskip
        \begin{subfigure}[b]{0.3\textwidth}
            \centering
            \fbox{\includegraphics[width=\dimexpr\textwidth-2\fboxsep-2\fboxrule\relax]{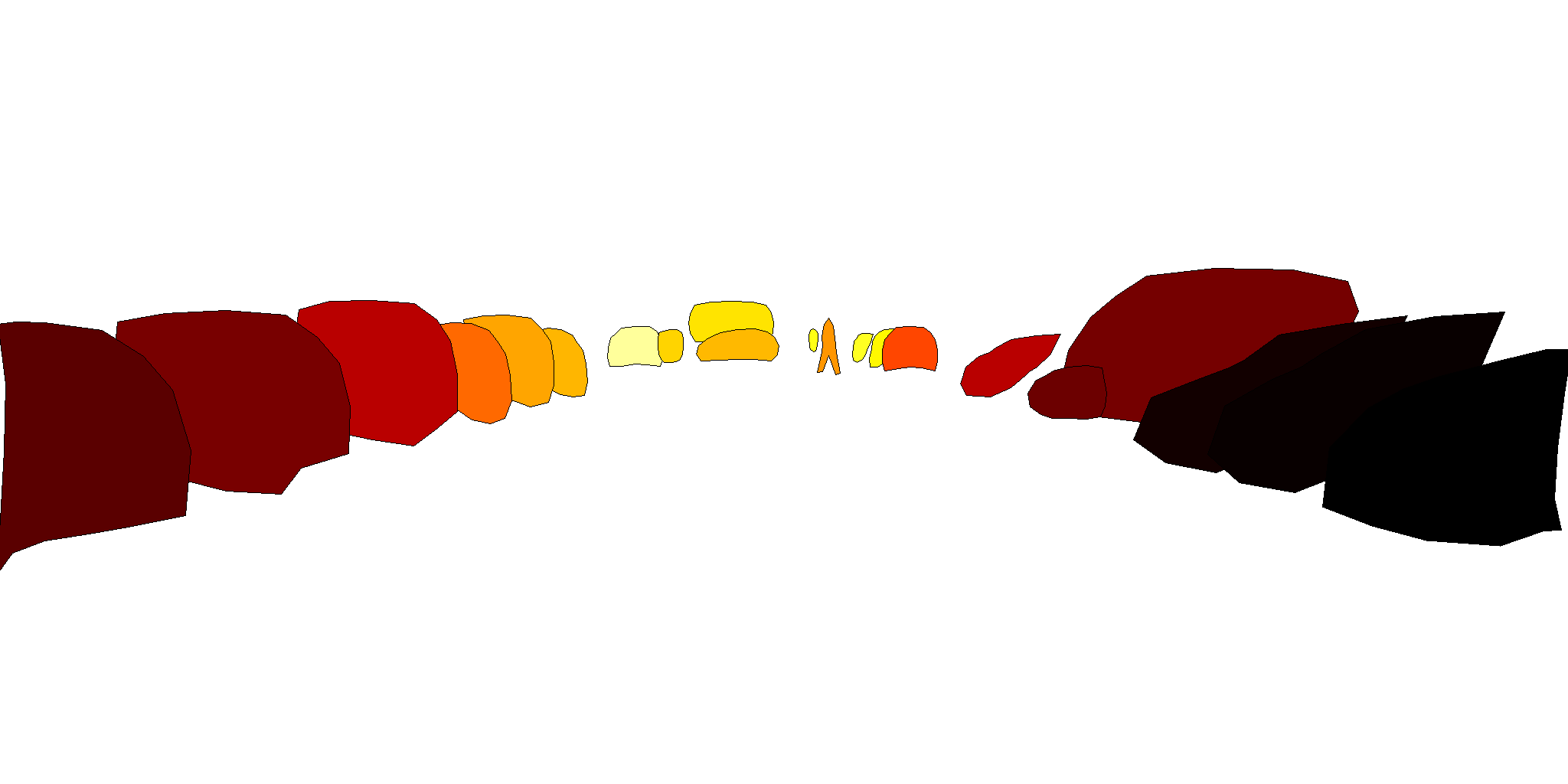}}
        \end{subfigure}
        \hspace{0.5em}
        \begin{subfigure}[b]{0.3\textwidth}  
            \centering 
            \fbox{\includegraphics[width=\dimexpr\textwidth-2\fboxsep-2\fboxrule\relax]{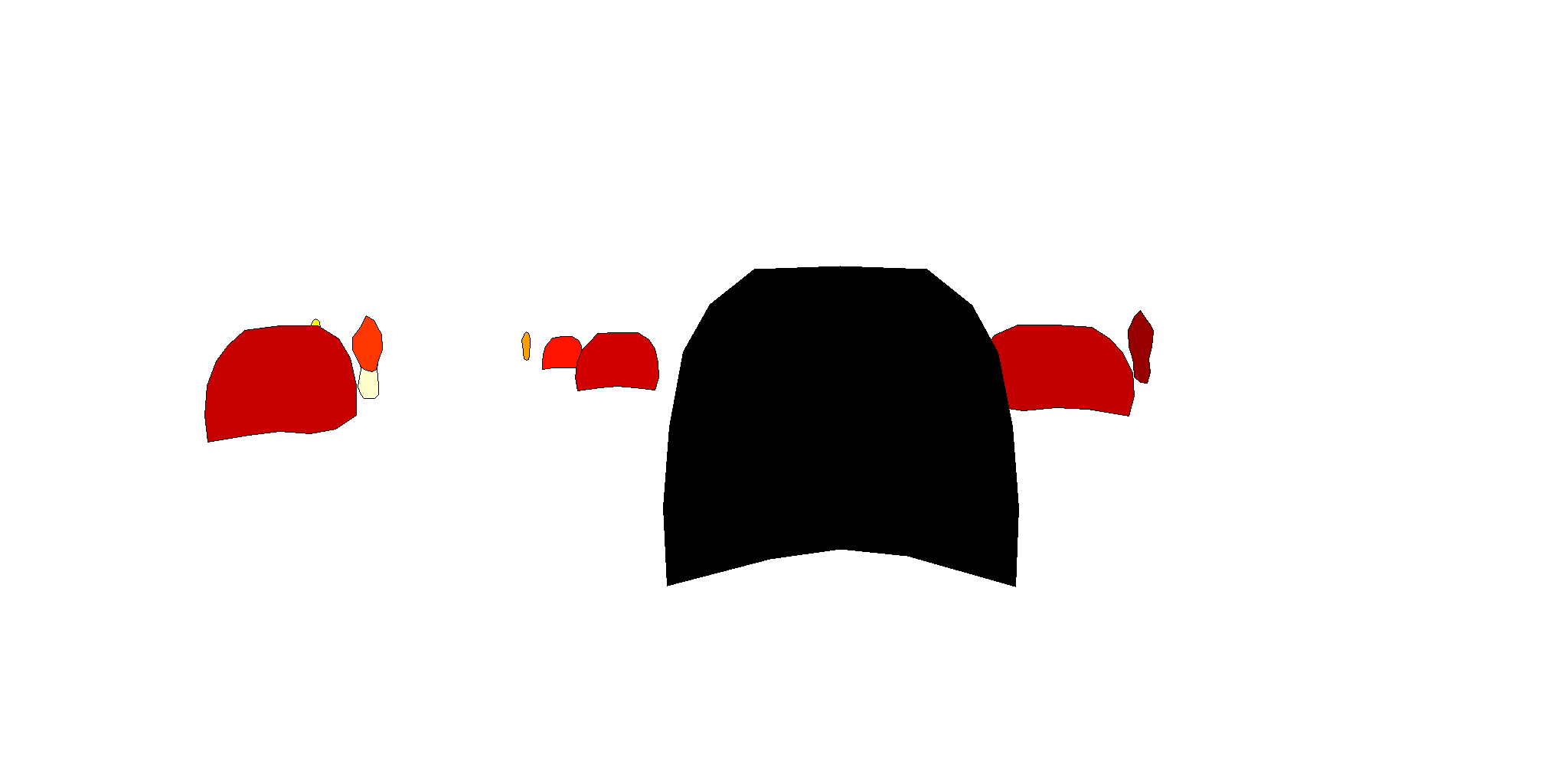}}
        \end{subfigure}
        \hspace{0.5em}
        \begin{subfigure}[b]{0.3\textwidth}
            \centering
            \fbox{\includegraphics[width=\dimexpr\textwidth-2\fboxsep-2\fboxrule\relax]{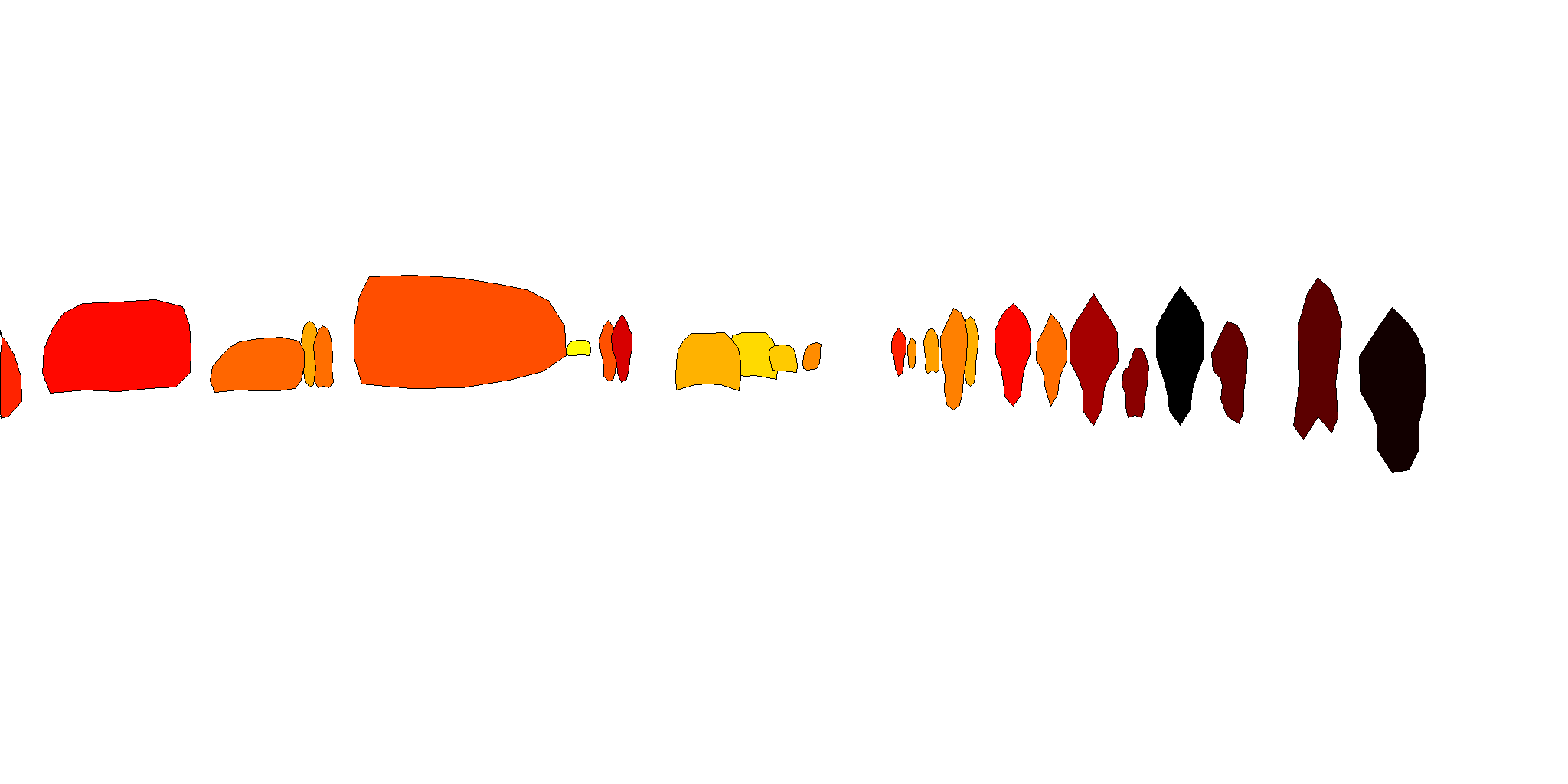}}
        \end{subfigure}
        \caption{Qualitative instance segmentation results of CenterPoly and their corresponding relative depth estimation taken from the Cityscapes test set. For the relative depth, darker is closer, and lighter is further.} 
        \label{qualitative}
        \vspace{-1.2em}

    \end{figure*}

We present some qualitative results in figure~\ref{qualitative}, in which we can notice that CenterPoly can accurately segment the legs of pedestrians, even for very small ones. Also, separating dense instances of very small objects can be done, showing that CenterPoly does indeed instance segmentation and not semantic segmentation. However, the segmented bicycles do show some limitations of using bounding polygons, as CenterPoly struggles to produce a very fine masks.

\section{Discussion}
\subsection{Ablation Study}
\label{secablation}

\begin{table*}
\caption{Ablation study of different parts of CenterPoly as well as performance with various backbones. AP and AP50\% on the validation set of Cityscapes. Depth refer to the use of our relative depth map. Elliptical refers to the use of elliptical GT. Center of gravity refer to the center heatmaps being at the center of gravity instead of the center of the bounding boxes. }
\begin{center}
\begin{tabular}{|c|c|c|c|c|c|c|c|c|}
\hline
Backbone & Depth & Elliptical & C. of Grav. & Nbr. Vertices & Res. & AP & AP50\% & Runtime(s) \\
\hline\hline
Hourglass & \cmark & \cmark & \cmark &  32 & 1024$\times$512 & 18.4 & 46.0 & 0.047 \\

Hourglass & \cmark & \cmark & \cmark &  16 & 1024$\times$512 & \textbf{18.5} & \textbf{46.2} & 0.046\\
Hourglass & \cmark & \cmark & \cmark &  8 & 1024$\times$512 & 15.6 & 42.6 & 0.046 \\
Hourglass & \cmark & \cmark & \xmark &  16 & 1024$\times$512 & 17.5 & 44.7 & 0.046\\
Hourglass & \cmark & \xmark & \cmark &  16 & 1024$\times$512 & 17.4 & 43.0 & 0.046\\
Hourglass & \xmark & \cmark & \cmark & 16 &1024$\times$512 & 17.8 & 45.4 & 0.046\\
Hourglass & \cmark & \cmark & \cmark & 16 &512$\times$512 & 10.8 &  25.3 & \textbf{0.026}\\
\hline
DCNv2 ResNet-50 & \cmark & \cmark & \cmark & 16 &1024$\times$512 & \textbf{15.4} & 36,9 & 0.023\\
DCNv2 ResNet-50 & \xmark & \cmark & \cmark & 16 &1024$\times$512 & 14.9 & \textbf{38.1} & 0.023\\
DCNv2 ResNet-50 & \cmark & \cmark & \cmark & 16 &512$\times$512 & 9.2 & 22.0 & \textbf{0.016}\\
\hline
DCNv2 ResNet-18 & \cmark & \cmark & \cmark & 16 &1024$\times$512 & \textbf{8.9} & \textbf{23.3} & 0.016\\
DCNv2 ResNet-18 & \xmark & \cmark & \cmark & 16 &1024$\times$512 & 8.7 & 23.2 & 0.016\\
DCNv2 ResNet-18 & \cmark & \cmark & \cmark & 16 &512$\times$512 & 5.2 & 13.3 & \textbf{0.01}\\
\hline
\end{tabular}
\end{center}
\label{ablation}
\vspace{0.5em}
\end{table*}

In order to evaluate the possible speed / accuracy trade-offs, we trained our method using various, lighter backbones, namely DCNv2~\cite{zhu2019deformable} with ResNet-18~\cite{he2016deep_resnet} and ResNet-50~\cite{he2016deep_resnet}. We also trained with various configurations to assess and demonstrate the benefits of our contributions. We performed our ablation study on the validation set of Cityscapes, and show our results in table~\ref{ablation}. We did not use the test set, as evaluation on the server is restricted.  

From the ablation study results, we can conclude the following. Firstly, we can notice that using the depth map is very useful for CenterPoly with all the tested backbones, allowing the network to better attribute pixels of overlapping instances. Secondly, the AP seems to plateau when more polygon vertices are used. This shows that a very large number of vertices is not necessary to capture the external contour of objects. This is why we chose to use 16 vertices in our final results. Thirdly, we can observe a decrease in AP when not using our elliptical center heatmaps and our center of gravity locations, showing their benefit for localizing objects more accurately. Elliptical center heatmaps help in the case of long rectangular objects, as the loss penalizes less an error over the long axis than the short one. Using the center of gravity helps CenterPoly to regress the polygon vertices as center offsets, contrarily to CenterNet that regresses a width and height. Fourthly, the best speed / accuracy trade-off comes from using a lighter backbone rather than reducing the resolution, as it makes the detection and segmentation of small instances harder. The DCNv2 ResNet-50 is a very interesting option at double the speed of the Hourglass with only 3 AP less. 

However, instance segmentation is a very complex task, and we can see that using lighter backbones also reduces the performance. We can conclude that we require a certain level of expressivity in our backbone network to maintain a good performance.  
Finally, the number of polygon vertices (from 8 to 32) does not affect significantly the speed of our method.

\subsection{Limitations}
One of the limitations of our method for instance segmentation is of course the lack of detail of the obtained masks: the precision of our masks is limited by the number of vertices of the polygons that we use, as well as because they cannot include holes. Still, we can see in figure~\ref{qualitative} that our method has enough precision to segment the legs of pedestrians. Nevertheless, this limitation explains the fact that we rank better in AP50 than in AP, which is a very strict metric. 

Another limitation is that due to training and network design, a specific trained model will always output the same number of vertices for every polygon, even if it does not need to. There might be some future work to do in that direction to try to further speed up the method by using fewer vertices when not needed. Poly-YOLO tackles this problem by introducing a confidence score for each vertex, but this is not a direction we wanted to take as it introduces yet another value. It is also counterproductive in our case as we wanted to reduce the number of values needed to represent the polygons. 

\section{Conclusion}
In this work, we presented CenterPoly, a novel real-time instance segmentation method that segments objects using bounding polygons. We designed a policy to select ground-truth vertices, which helps the training process. We show the importance of depth map to improve the accuracy of the polygons because of the overlap problem. CenterPoly ranks first among real-time instance segmentation methods on the popular Cityscapes, KITTI and IDD datasets. 

\section{Acknowledgment}
\vspace{-0.2em}
We acknowledge the support of the Natural Sciences and Engineering  Research Council of Canada (NSERC), [RDCPJ 508883 - 17], and the support of Genetec.
{\small
\bibliographystyle{ieee_fullname}
\bibliography{bib}

\begin{thebibliography}{10}\itemsep=-1pt

\bibitem{acuna2018efficientpolygonrnn}
David Acuna, Huan Ling, Amlan Kar, and Sanja Fidler.
\newblock Efficient interactive annotation of segmentation datasets with
  polygon-rnn++.
\newblock In {\em Proceedings of the IEEE conference on Computer Vision and
  Pattern Recognition}, pages 859--868, 2018.

\bibitem{Alhaija2018IJCV_kitti}
Hassan Alhaija, Siva Mustikovela, Lars Mescheder, Andreas Geiger, and Carsten
  Rother.
\newblock Augmented reality meets computer vision: Efficient data generation
  for urban driving scenes.
\newblock {\em International Journal of Computer Vision (IJCV)}, 2018.

\bibitem{bolya2019yolact}
Daniel Bolya, Chong Zhou, Fanyi Xiao, and Yong~Jae Lee.
\newblock Yolact: Real-time instance segmentation.
\newblock In {\em Proceedings of the IEEE/CVF International Conference on
  Computer Vision}, pages 9157--9166, 2019.

\bibitem{chen2021edge}
Chen Chen, Bin Liu, Shaohua Wan, Peng Qiao, and Qingqi Pei.
\newblock An edge traffic flow detection scheme based on deep learning in an
  intelligent transportation system.
\newblock {\em IEEE Transactions on Intelligent Transportation Systems},
  22(3):1840--1852, 2021.

\bibitem{cordts2016Cityscapes}
Marius Cordts, Mohamed Omran, Sebastian Ramos, Timo Rehfeld, Markus Enzweiler,
  Rodrigo Benenson, Uwe Franke, Stefan Roth, and Bernt Schiele.
\newblock The cityscapes dataset for semantic urban scene understanding.
\newblock In {\em Proceedings of the IEEE conference on Computer Vision and
  Pattern Recognition}, pages 3213--3223, 2016.

\bibitem{duan2019centernet}
Kaiwen Duan, Song Bai, Lingxi Xie, Honggang Qi, Qingming Huang, and Qi Tian.
\newblock Centernet: Keypoint triplets for object detection.
\newblock In {\em Proceedings of the IEEE International Conference on Computer
  Vision}, pages 6569--6578, 2019.

\bibitem{fu2019retinamask}
Cheng-Yang Fu, Mykhailo Shvets, and Alexander~C Berg.
\newblock Retinamask: Learning to predict masks improves state-of-the-art
  single-shot detection for free.
\newblock {\em arXiv preprint arXiv:1901.03353}, 2019.

\bibitem{rcnn_Girshick_2014_CVPR}
Ross Girshick, Jeff Donahue, Trevor Darrell, and Jitendra Malik.
\newblock Rich feature hierarchies for accurate object detection and semantic
  segmentation.
\newblock In {\em Proceedings of the IEEE conference on Computer Vision and
  Pattern Recognition}, pages 580--587, 2014.

\bibitem{he2017maskrcnn}
Kaiming He, Georgia Gkioxari, Piotr Doll{\'a}r, and Ross Girshick.
\newblock Mask r-cnn.
\newblock In {\em Proceedings of the IEEE international conference on computer
  vision}, pages 2961--2969, 2017.

\bibitem{he2016deep_resnet}
Kaiming He, Xiangyu Zhang, Shaoqing Ren, and Jian Sun.
\newblock Deep residual learning for image recognition.
\newblock In {\em Proceedings of the IEEE conference on Computer Vision and
  Pattern Recognition}, pages 770--778, 2016.

\bibitem{hsu2018learningLCIS}
Yen-Chang Hsu, Zheng Xu, Zsolt Kira, and Jiawei Huang.
\newblock Learning to cluster for proposal-free instance segmentation.
\newblock In {\em 2018 International Joint Conference on Neural Networks
  (IJCNN)}, pages 1--8. IEEE, 2018.

\bibitem{hurtik2020poly-yolo}
Petr Hurtik, Vojtech Molek, Jan Hula, Marek Vajgl, Pavel Vlasanek, and Tomas
  Nejezchleba.
\newblock Poly-yolo: higher speed, more precise detection and instance
  segmentation for yolov3.
\newblock {\em arXiv preprint arXiv:2005.13243}, 2020.

\bibitem{kingma2014adam}
Diederik~P Kingma and Jimmy Ba.
\newblock Adam: A method for stochastic optimization.
\newblock {\em arXiv preprint arXiv:1412.6980}, 2014.

\bibitem{kirillov2017instancecut}
Alexander Kirillov, Evgeny Levinkov, Bjoern Andres, Bogdan Savchynskyy, and
  Carsten Rother.
\newblock Instancecut: from edges to instances with multicut.
\newblock In {\em Proceedings of the IEEE Conference on Computer Vision and
  Pattern Recognition}, pages 5008--5017, 2017.

\bibitem{law2018cornernet}
Hei Law and Jia Deng.
\newblock Cornernet: Detecting objects as paired keypoints.
\newblock In {\em Proceedings of the European Conference on Computer Vision
  (ECCV)}, pages 734--750, 2018.

\bibitem{Lin_2017_CVPR_FPN}
Tsung-Yi Lin, Piotr Doll{\'a}r, Ross Girshick, Kaiming He, Bharath Hariharan,
  and Serge Belongie.
\newblock Feature pyramid networks for object detection.
\newblock In {\em Proceedings of the IEEE conference on Computer Vision and
  Pattern Recognition}, pages 2117--2125, 2017.

\bibitem{lin2018focal_retina}
Tsung-Yi Lin, Priyal Goyal, Ross Girshick, Kaiming He, and Piotr Doll{\'a}r.
\newblock Focal loss for dense object detection.
\newblock {\em IEEE transactions on pattern analysis and machine intelligence},
  2018.

\bibitem{lin2014microsoft}
Tsung-Yi Lin, Michael Maire, Serge Belongie, James Hays, Pietro Perona, Deva
  Ramanan, Piotr Doll{\'a}r, and C~Lawrence Zitnick.
\newblock Microsoft coco: Common objects in context.
\newblock In {\em European conference on computer vision}, pages 740--755.
  Springer, 2014.

\bibitem{liu2018path_panet}
Shu Liu, Lu Qi, Haifang Qin, Jianping Shi, and Jiaya Jia.
\newblock Path aggregation network for instance segmentation.
\newblock In {\em Proceedings of the IEEE conference on Computer Vision and
  Pattern Recognition}, pages 8759--8768, 2018.

\bibitem{liu2016ssd}
Wei Liu, Dragomir Anguelov, Dumitru Erhan, Christian Szegedy, Scott Reed,
  Cheng-Yang Fu, and Alexander~C Berg.
\newblock Ssd: Single shot multibox detector.
\newblock In {\em European conference on computer vision}, pages 21--37.
  Springer, 2016.

\bibitem{Mazzini_2019_CVPR_Workshops}
Davide Mazzini and Raimondo Schettini.
\newblock Spatial sampling network for fast scene understanding.
\newblock In {\em Proceedings of the IEEE/CVF Conference on Computer Vision and
  Pattern Recognition (CVPR) Workshops}, June 2019.

\bibitem{newell2016stacked}
Alejandro Newell, Kaiyu Yang, and Jia Deng.
\newblock Stacked hourglass networks for human pose estimation.
\newblock In {\em European conference on computer vision}, pages 483--499.
  Springer, 2016.

\bibitem{NEURIPS2019_9015_pytorch}
Adam Paszke, Sam Gross, Francisco Massa, Adam Lerer, James Bradbury, Gregory
  Chanan, Trevor Killeen, Zeming Lin, Natalia Gimelshein, Luca Antiga, Alban
  Desmaison, Andreas Kopf, Edward Yang, Zachary DeVito, Martin Raison, Alykhan
  Tejani, Sasank Chilamkurthy, Benoit Steiner, Lu Fang, Junjie Bai, and Soumith
  Chintala.
\newblock Pytorch: An imperative style, high-performance deep learning library.
\newblock In H. Wallach, H. Larochelle, A. Beygelzimer, F. d\textquotesingle
  Alch\'{e}-Buc, E. Fox, and R. Garnett, editors, {\em Advances in Neural
  Information Processing Systems 32}, pages 8024--8035. Curran Associates,
  Inc., 2019.

\bibitem{peng2020deepsnake}
Sida Peng, Wen Jiang, Huaijin Pi, Xiuli Li, Hujun Bao, and Xiaowei Zhou.
\newblock Deep snake for real-time instance segmentation.
\newblock In {\em Proceedings of the IEEE/CVF Conference on Computer Vision and
  Pattern Recognition}, pages 8533--8542, 2020.

\bibitem{perreault2020spotnet}
Hughes Perreault, Guillaume-Alexandre Bilodeau, Nicolas Saunier, and Maguelonne
  H{\'e}ritier.
\newblock Spotnet: Self-attention multi-task network for object detection.
\newblock In {\em 2020 17th Conference on Computer and Robot Vision (CRV)},
  pages 230--237. IEEE, 2020.

\bibitem{yolo_redmon2016you}
Joseph Redmon, Santosh Divvala, Ross Girshick, and Ali Farhadi.
\newblock You only look once: Unified, real-time object detection.
\newblock In {\em Proceedings of the IEEE conference on Computer Vision and
  Pattern Recognition}, pages 779--788, 2016.

\bibitem{redmon2017yolo9000}
Joseph Redmon and Ali Farhadi.
\newblock Yolo9000: better, faster, stronger.
\newblock In {\em Proceedings of the IEEE conference on Computer Vision and
  Pattern Recognition}, pages 7263--7271, 2017.

\bibitem{redmon2018yolov3}
Joseph Redmon and Ali Farhadi.
\newblock Yolov3: An incremental improvement.
\newblock {\em arXiv preprint arXiv:1804.02767}, 2018.

\bibitem{ren2017endrecattend}
Mengye Ren and Richard~S Zemel.
\newblock End-to-end instance segmentation with recurrent attention.
\newblock In {\em Proceedings of the IEEE conference on Computer Vision and
  Pattern Recognition}, pages 6656--6664, 2017.

\bibitem{ren2015faster}
Shaoqing Ren, Kaiming He, Ross Girshick, and Jian Sun.
\newblock Faster r-cnn: Towards real-time object detection with region proposal
  networks.
\newblock In {\em Advances in neural information processing systems}, pages
  91--99, 2015.

\bibitem{UB18}
J. Uhrig, E. Rehder, B. Fr{\"o}hlich, U. Franke, and T. Brox.
\newblock Box2pix: Single-shot instance segmentation by assigning pixels to
  object boxes.
\newblock In {\em IEEE Intelligent Vehicles Symposium (IV)}, 2018.

\bibitem{varma2019idd}
Girish Varma, Anbumani Subramanian, Anoop Namboodiri, Manmohan Chandraker, and
  CV Jawahar.
\newblock Idd: A dataset for exploring problems of autonomous navigation in
  unconstrained environments.
\newblock In {\em 2019 IEEE Winter Conference on Applications of Computer
  Vision (WACV)}, pages 1743--1751. IEEE, 2019.

\bibitem{wang2020solo}
Xinlong Wang, Tao Kong, Chunhua Shen, Yuning Jiang, and Lei Li.
\newblock Solo: Segmenting objects by locations.
\newblock In {\em European Conference on Computer Vision}, pages 649--665.
  Springer, 2020.

\bibitem{wang2020solov2}
Xinlong Wang, Rufeng Zhang, Tao Kong, Lei Li, and Chunhua Shen.
\newblock Solov2: Dynamic and fast instance segmentation.
\newblock {\em Advances in Neural Information Processing Systems}, 2020.

\bibitem{xie2020polarmask}
Enze Xie, Peize Sun, Xiaoge Song, Wenhai Wang, Xuebo Liu, Ding Liang, Chunhua
  Shen, and Ping Luo.
\newblock Polarmask: Single shot instance segmentation with polar
  representation.
\newblock In {\em Proceedings of the IEEE/CVF conference on Computer Vision and
  Pattern Recognition}, pages 12193--12202, 2020.

\bibitem{eseseg_xu2019explicit}
Wenqiang Xu, Haiyang Wang, Fubo Qi, and Cewu Lu.
\newblock Explicit shape encoding for real-time instance segmentation.
\newblock In {\em Proceedings of the IEEE/CVF International Conference on
  Computer Vision}, pages 5168--5177, 2019.

\bibitem{zhou2021simple_unidet}
Xingyi Zhou, Vladlen Koltun, and Philipp Kr{\"a}henb{\"u}hl.
\newblock Simple multi-dataset detection.
\newblock {\em arXiv preprint arXiv:2102.13086}, 2021.

\bibitem{centernet_zhou2019objects}
Xingyi Zhou, Dequan Wang, and Philipp Kr{\"a}henb{\"u}hl.
\newblock Objects as points.
\newblock {\em arXiv preprint arXiv:1904.07850}, 2019.

\bibitem{zhu2019deformable}
Xizhou Zhu, Han Hu, Stephen Lin, and Jifeng Dai.
\newblock Deformable convnets v2: More deformable, better results.
\newblock In {\em Proceedings of the IEEE/CVF Conference on Computer Vision and
  Pattern Recognition}, pages 9308--9316, 2019.

\end{thebibliography}
}

\end{document}